\documentclass[10pt,twocolumn,letterpaper]{article}

\usepackage[pagenumbers]{cvpr}  

\usepackage{graphicx}
\usepackage[abs]{overpic}
\usepackage{capt-of}
\usepackage{xcolor}
\usepackage{amsmath,amssymb,amsfonts,dsfont,pifont,bm,bbm,mathrsfs,mathtools,nicefrac}
\usepackage{algorithm,algpseudocode,listings}
\usepackage{booktabs,multirow,adjustbox,diagbox,threeparttable}
\definecolor{citeblue}{RGB}{48,111,186}
\usepackage[pagebackref,breaklinks,colorlinks,citecolor=citeblue,bookmarks=false]{hyperref}
\usepackage[capitalize]{cleveref}  

\crefname{section}{Sec.}{Secs.}
\Crefname{section}{Section}{Sections}
\crefname{table}{Tab.}{Tabs.}
\Crefname{table}{Table}{Tables}
\crefname{figure}{Fig.}{Figs.}
\Crefname{figure}{Figure}{Figures}
\crefname{equation}{Eq.}{Eqs.}
\Crefname{equation}{Equation}{Equations}
\newcommand{\G}{\mathsf{G}}
\newcommand{\C}{\mathsf{C}}

\newcommand{\V}{V}
\newcommand{\Vdens}{\V^{\mathrm{Density}}}
\newcommand{\Vfeat}{\V^{\mathrm{RGB}}}
\newcommand{\VLBS}{\V^{\mathrm{LBS}}}
\newcommand{\f}{F}
\newcommand{\T}{T}
\newcommand{\Tp}{T_p}
\newcommand{\rot}{R}
\newcommand{\rotp}{R_p}
\newcommand{\tr}{t}
\newcommand{\trp}{t_p}
\newcommand{\kpthree}{K^{3D}}
\newcommand{\kptwo}{K^{2D}}
\newcommand{\nk}{N_k}

\newcommand{\id}{\alpha}
\newcommand{\nf}{N_f}

\newcommand{\nemb}{N_e}
\newcommand{\emb}{\mathbf{e}}

\newcommand{\np}{N_p}
\newcommand{\voxsize}{S}

\newcommand{\yaw}{\Theta}
\newcommand{\shapecode}{c^s}
\newcommand{\posecode}{c^p}
\newcommand{\shapecodelen}{N_{\shapecode}}
\newcommand{\posecodelen}{N_{\posecode}}

\newcommand{\secpostvspace}{\vspace{-0.05cm}}
\newcommand{\secprevspace}{\vspace{-0.05cm}}

\newcommand{\apppostvspace}{\vspace{-0.2cm}}
\newcommand{\appprevspace}{\vspace{-0.2cm}}
\newcommand{\R}{\mathbb{R}}
\newcommand{\N}{\mathbb{N}}

\newcommand{\figref}[1]{Fig.~\ref{#1}}
\newcommand{\apref}[1]{Appx~\ref{#1}}
\newcommand{\secref}[1]{Sec.~\ref{#1}}

\newcommand\blfootnote[1]{%
  \begingroup
  \renewcommand\thefootnote{}\footnote{#1}%
  \addtocounter{footnote}{-1}%
  \endgroup
}

\begin{document}

\title{Unsupervised Volumetric Animation}

\twocolumn[{%
\author{
Aliaksandr Siarohin\\
\normalsize Snap Inc.
\and
Willi Menapace$^*$\\
\normalsize University of Trento 
\and
Ivan Skorokhodov$^*$\\
\normalsize KAUST \\
\and
Kyle Olszewski\\
\normalsize Snap Inc.
\and
Jian Ren\\
\normalsize Snap Inc.
\and
Hsin-Ying Lee\\
\normalsize Snap Inc.
\and
Menglei Chai\\
\normalsize Snap Inc.
\and
Sergey Tulyakov\\
\normalsize Snap Inc.
\vspace{-0.5cm}
}

\maketitle
\renewcommand\twocolumn[1][]{#1}
    
    \centering 
    \includegraphics[width=\linewidth]{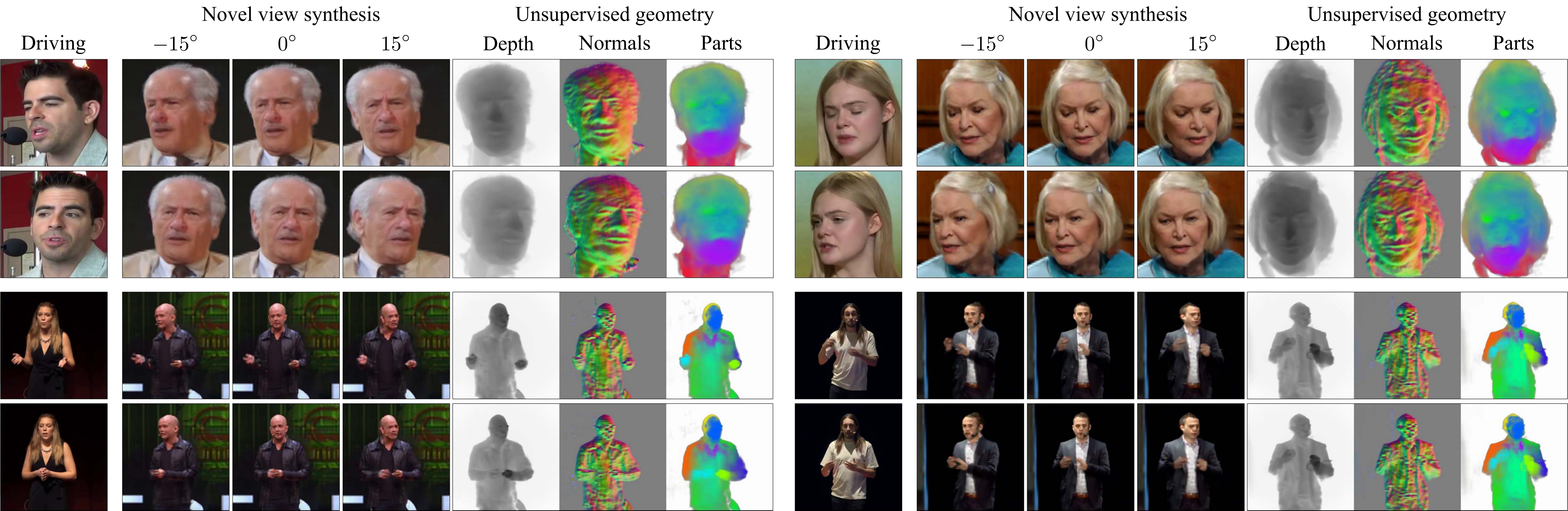}
    \captionof{figure}{
    \textbf{Unsupervised Volumetric Animation (UVA).} Selected animation results for faces and bodies. Given a driving image sequence and a source image (not shown), UVA renders realistic animations and \emph{simultaneously} generates novel views of the animated object. With our reconstruction loss, our method also generates high-fidelity depth and normals, and identifies semantically meaningful object parts. 
    } \label{fig:teaser}
    \vspace{3mm}
}]

\maketitle

\begin{abstract}

We propose a novel approach for unsupervised 3D animation of non-rigid deformable objects. Our method learns the 3D structure and dynamics of objects solely from single-view RGB videos, and can decompose them into semantically meaningful parts that can be tracked and animated. Using a 3D autodecoder framework, paired with a keypoint estimator via a differentiable PnP algorithm, our model learns the underlying object geometry and parts decomposition in an entirely unsupervised manner. This allows it to perform 3D segmentation, 3D keypoint estimation, novel view synthesis, and animation. We primarily evaluate the framework on two video datasets: VoxCeleb $256^2$ and TEDXPeople $256^2$. In addition, on the Cats $256^2$ image dataset, we show it even learns compelling 3D geometry from still images. Finally, we show our model can obtain animatable 3D objects from a single or few images~\footnote{Code and visual results available on our \href{https://snap-research.github.io/unsupervised-volumetric-animation}{project website}.}\blfootnote{\hspace{-0.21cm} $^*$ Work done while interning at Snap.}.
\end{abstract}
\vspace{-0.3cm}

\secprevspace
\section{Introduction}
\secpostvspace
The ability to realistically animate a dynamic object seen in a single image enables compelling creative tasks. Such applications range from tractable and cost-effective approaches to visual effects for cinema and television, to more lightweight consumer applications (\eg, enabling arbitrary users to create ``performances'' by famous modern or historical figures).
However, this requires understanding the object's structure and motion patterns from a single static depiction.
Efforts in this field are primarily divided into two approaches: those that outsource this understanding to existing, off-the-shelf models specific to an object category that capture its particular factors of variation; and those that learn the object structure from the raw training data itself. The former group employs \emph{supervision}, and thus requires knowledge about the animated object (\eg, the plausible range of shapes and motions of human faces or bodies). The latter group is \emph{unsupervised}, providing the flexibility needed for a wider range of arbitrary object categories.

Significant progress has been made recently in the domain of unsupervised image animation. Methods in this category typically learn a motion model based on object parts and the corresponding transformations applied to them. Initially, such transformations were modeled using a simple set of sparse keypoints. Further works improved the motion representation~\cite{FOMM,MRAA}, learned latent motion dictionaries~\cite{LatentImageAnimator}, kinematic chains~\cite{kinematictree} or used thin-plate spline transformations~\cite{TPSMotion}. However, broadly speaking, all such works propose \emph{2D motion representations}, warping the pixels or features of the input image such that they correspond to the pose of a given driving image. As such, prior unsupervised animation methods offer means to perform 2D animation only, and are inherently limited in modeling complex, 3D effects, such as occlusions, viewpoint changes, and extreme rotations, which can only be explained and addressed appropriately when considering the 3D nature of the observed objects. 

Our work fundamentally differs from prior 2D works in that it is the first to explore unsupervised image animation in 3D. This setting is substantially more challenging compared to classical 2D animation for several reasons. First, as the predicted regions or parts now exist in a 3D space, it is quite challenging to identify and plausibly control them from only 2D videos without extra supervision. Second, this challenge is further compounded by the need to properly model the distribution of the camera in 3D, which is a problem in its own right~\cite{CAMPARI}, with multiple 3D generators resorting to existing pose predictors to facilitate the learning of the underlying 3D geometry~\cite{EG3D,EpiGRAF}.
Finally, in 3D space, there exists no obvious and tractable counterpart for the bias of 2D CNNs, which are essential for unsupervised keypoint detection frameworks for 2D images~\cite{MonkeyNet}.

We offer a solution to these challenges. Our framework maps an embedding of each object to a canonical volumetric representation, parameterized with a voxel grid, containing volumetric density and appearance. To allow for non-rigid deformations of the canonical object representation, we assume the object consists of a certain number of rigid parts which are softly assigned to each of the points in the canonical volume. A procedure based on linear blend skinning is employed to produce the deformed volume according to the pose of each part. Rather than directly estimating the poses, we introduce a set of learnable 3D canonical keypoints for each part, and leverage the 2D inductive bias of 2D CNNs to predict a set of corresponding 2D keypoints in the current frame. We propose the use of a differentiable Perspective-n-Point (PnP) algorithm to estimate the corresponding pose, explicitly linking 2D observations to our 3D representation. This framework allows us to propagate the knowledge from 2D images to our 3D representation, thereby learning rich and detailed geometry for diverse object categories using a photometric reconstruction loss as our driving objective. The parts are learned in an unsupervised manner, yet they converge to meaningful volumetric object constituents. For example, for faces, they correspond to the jaw, hair, neck, and the left and right eyes and cheeks. For bodies, the same approach learns parts to represent the torso, head, and each hand. Examples of these parts are given in Fig.~\ref{fig:teaser}.

To simplify the optimization, we introduce a two-stage strategy, in which we start by learning a single part such that the overall geometry is learned, and proceed by allowing the model to discover the remaining parts so that animation is possible. When the object is represented with a single part, the model can perform 3D reconstruction and novel view synthesis. When more parts are used, our method allows us to not only identify meaningful object parts, but to perform non-rigid animation \emph{and} novel view synthesis \emph{at the same time}. Examples of images animated using our Unsupervised Volumetric Animation (UVA) are given in Fig.~\ref{fig:teaser}.

We train our framework on three diverse datasets containing images or videos of various objects. We first show that our method learns meaningful 3D geometry when trained on still images of cat faces~\cite{Cats_dataset}. We then train our method on the VoxCeleb~\cite{voxceleb} and TEDXPeople\cite{stylepeople} video datasets to evaluate 3D animation. Since our method is the first to consider unsupervised 3D animation, we further introduce evaluation metrics assessing novel view synthesis and animation quality when only single-view data is available. 
\secprevspace
\section{Related work}
\secpostvspace
\noindent\textbf{3D-aware image and video synthesis} experienced substantial progress over the last two years.
Early works~\cite{GRAF, Giraffe, CAMPARI} used Neural Radiance Fields (NeRFs)~\cite{NeRF} as a 3D representation to synthesize simple objects and often considered synthetic datasets~\cite{GRAF, 3D-shapenets}.
They spurred a line of works that scaled the generator and increased its efficiency to attain high-resolution 3D synthesis~\cite{StyleNeRF, EG3D, EpiGRAF, GIRAFFE-HD, lolnerf}.
These works rely on different types of volumetric representations such as a coordinate-MLP~\cite{piGAN}, voxel-grids~\cite{BlockGAN}, tri-planes~\cite{EG3D,EpiGRAF}, generative manifolds~\cite{GRAM}, multi-plane representations~\cite{GMPI}, and signed distance functions~\cite{StyleSDF}.
Further works combined implicit video synthesis~\cite{DIGAN, StyleGAN-V} techniques with that of volumetric rendering~\cite{StyleNeRF} to generate 3D-aware videos~\cite{3D_video_gen}.
A common requirement of these methods is access to the ground truth camera distribution (\eg,~\cite{piGAN, StyleNeRF, GRAF, MVCGAN, lolnerf}) or even the known camera poses for each training image~\cite{EG3D, GMPI, EpiGRAF, GRAM}.
This gives a strong inductive bias towards recovering the proper 3D geometry~\cite{GMPI, EpiGRAF}.
Our work shows that it is possible to learn rich geometry and object parts decomposition in a completely unsupervised manner in a non-adversarial framework.

\noindent\textbf{Unsupervised 3D reconstruction.} Unsupervised reconstruction of 3D objects from image or video collections is a long standing problem~\cite{CategoryMesh,ProbablySymetric,ConsistentMesh, dove, lasr, viser, banmo}. Initial attempts~\cite{CategoryMesh} utilize image collections and try to predict camera, mesh displacement parameters and texture, render the mesh and use reconstruction as the main guiding signal. Later work~\cite{ProbablySymetric} proposes to further improve this pipeline by incorporating additional knowledge about object symmetry. However, those works did not model deformations, which was addressed later in works~\cite{banmo, viser, lasr, ConsistentMesh, dove} that propose to train on video datasets. Most of them~\cite{banmo, viser, lasr, ConsistentMesh} optimize the object parameters for each frame, and thus can not be trained on a large dataset of videos. On the contrary, Dove~\cite{dove} infers the articulation parameters from individual frames, which allows training on large video dataset. However, Dove~\cite{dove} is a mesh based method, thus rendering quality is limited. Moreover, all of these methods utilize additional annotations such as template shapes~\cite{ConsistentMesh}, camera poses~\cite{banmo}, 2D keypoints~\cite{banmo}, optical flow~\cite{banmo, lasr, viser} or ground truth object masks~\cite{banmo, viser, lasr, ConsistentMesh, dove}. Instead, in our method everything, including object masks, was obtained in an purely unsupervised way from video data only.

\noindent\textbf{Supervised image animation} requires an off-the-shelf keypoint predictor~\cite{FewShotZakharov,BiLayerZakharov,FewShotVid2Vid} or a 3D morphable model (3DMM) estimator~\cite{3DGuidedFineGrained,geng2020towards} run through the training dataset prior to training. To train such an estimator, one needs to have large amounts of labeled data for the task at a hand. 
Supervised animation works are typically designed for only one object category, such as bodies~\cite{LiquidWarpingGan, FewShotVid2Vid} or faces~\cite{FewShotZakharov}.
Among them, some support only a single object identity~\cite{EverybodyDanceNow}, others single- or few-shot cases~\cite{FewShotVid2Vid, fgTBNs}. 

Thanks to significant advances in neural rendering and 3D-aware synthesis, several works extended supervised animation to the 3D domain. Initially, a dataset with multiview videos was required to train animatable radiance fields~\cite{AnimatableNeRF}. Later, HumanNeRF~\cite{HumanNerf} and NeuMan~\cite{NeuMan} showed the feasibility of leveraging only a monocular video of the same subject. However, these models require fitting of a 3D model of human bodies to every frame of a video. With some exceptions~\cite{fgTBNs}, such methods do not support multiple identities with the same framework. In contrast, our method features a space in which all objects are represented in their canonical, animation-ready form.

\noindent\textbf{Unsupervised image animation} is the most related group of works to ours. These works do not require supervision beyond photometric reconstruction loss and, hence, support a variety of object categories with one framework~\cite{FOMM,MonkeyNet,ImplicitWarpingNVIDIA,MRAA}. A key focus area of such works is to design appropriate motion representations for animation~\cite{X2FACE, lorenz2019unsupervised, FOMM, MRAA}. A number of improved representations have been proposed, such as those setting additional constraints on a kinematic tree~\cite{kinematictree}, and thin-plate spline motion modelling~\cite{TPSMotion}.  A further work, titled Latent Image Animator~\cite{LatentImageAnimator}, learned a latent space for possible motions. Interestingly, a direction in the latent space is found to be responsible for generating novel views of the same subject. As we confirm experimentally, similarly to 2D image generators~\cite{StyleGAN2}, the direction cannot be reliably used to synthesize the novel views. Several recent works~\cite{VideoConferencing, MegaPortraits}, propose to use mixed schemes where pose of the object is supervised and expression is learned, such approaches works well for faces however did not generalize to other categories.


\secprevspace
\section{Method}
\label{sec:method}
\secpostvspace

\begin{figure*}
\centering
\includegraphics[width=\textwidth]{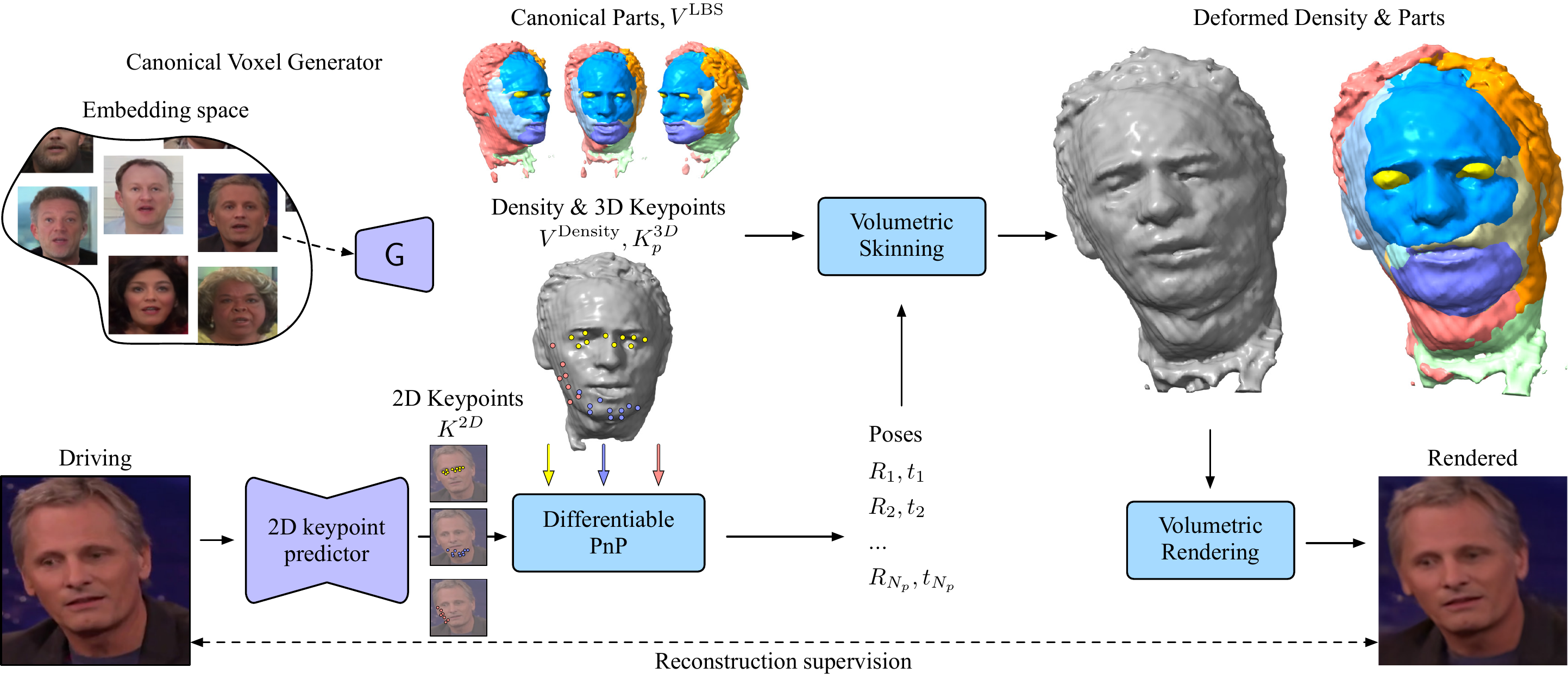}
\vspace{-0.7cm}
\caption{\textbf{Unsupervised Volumetric Animation} consists of a canonical voxel generator $\G$ mapping a point in the latent space to the canonical density, radiance and canonical parts. In the embedding space we show canonical shapes rendered under identity camera (faces have the same pose with mouth open). For each part, a set of canonical 3D keypoints $\kpthree$ is learnt during training. The 2D keypoint predictor uses a driving image to predict a set of 2D keypoints $\kptwo$, corresponding to $\kpthree$. The differentiable PnP algorithm is used to predict the pose of each part. Canonical density, radiance, poses and parts are then used to compute the deformed density and radiance via volumetric skinning. We then volumetrically render the deformed radiance to produce the rendered image. Note, that our approach does not use any knowledge about the object being animated, and is supervised using the reconstruction loss. Zoom-in for greater detail. 
\vspace{-0.5cm}
}
\label{fig:pipeline}
\end{figure*}


This section presents our method for unsupervised 3D animation of non-rigid deformable objects.
Our model trains on a set of images $\{\f_i, \id_i\}_{i=1}^{\nf}$, where $\f_i \in \R^{H \times W \times 3}$ is an image frame, $\id_i \in \N$ is an object identifier\footnote{We assume that we know which object instance appears in a video. In practice, this assumption is easily satisfied by assigning the same identity to all the frames of a given video.}, and $\nf$ is the number of frames in a video.
The primary training objective of our framework is the reconstruction task.
Given a frame $\f_i$ with identity $\id_i$, we reconstruct it using four core components (see \figref{fig:pipeline}).
First, Canonical Voxel Generator $\G$ maps a learnable identity-specific embedding $\emb \in \R^{\nemb}$ to an object's volumetric representation in the canonical pose, parametrized as a voxel grid.
Following~\cite{MonkeyNet, FOMM, MRAA}, we assume that each non-rigid object can be represented as a set of moving rigid parts.
In this way, our voxel generator segments the volume and assigns each 3D point to its corresponding object's part (\secref{sec:voxel_generator}).
Next, we define 2D keypoint predictor $\C$ with and the differentiable PnP~\cite{EPnP} algorithm to estimate each part's pose (position and orientation) in a given RGB frame $\f_i$ (\secref{sec:camera_predictor}). 
Subsequently, we employ a method based on linear blend skinning~\cite{lewis2000pose} to map the canonical object volume into a deformed one which represents the object in the current frame (\secref{sec:deformation_modeling}).
Finally, we use volumetric rendering~\cite{NeRF} to render the colors to the image space  (\secref{sec:volumetric_rendering}).

\secprevspace
\subsection{Canonical Voxel Generator}
\label{sec:voxel_generator}
\secpostvspace

We use a voxel grid $\V$ to parametrize the volume since we found it to provide the best trade-off between generation efficiency, expressivity and rendering speed.
Given an object's embedding $\emb \in \R^{\nemb}$, we use Canonical Voxel Generator $\G$ to produce a volume cube of size $\voxsize$:
\begin{equation}
\G(\emb) = \V = \left[  \Vdens \Vert \Vfeat \Vert \VLBS \right],
\end{equation}
where $\Vdens \in \R^{\voxsize^3}$ is the object's (discretized) density field in the canonical pose and $\Vfeat \in \R^{\voxsize^3 \times 3}$ is its (discretized) RGB radiance field.
To animate an object, we assume that it can be modeled as a set of rigid moving parts $p \in \{1, 2, ..., \np\}$~\cite{MonkeyNet, FOMM, MRAA}, so we use $\VLBS \in \R^{\voxsize^3 \times \np}$ to model a soft assignment of each point of the volume to one of the $\np$ parts.
Notably, we do not use any encoder to produce identity embeddings $\emb$ and instead optimize them directly during training~\cite{GLO}.
Examples of canonical density, parts, and rendered canonical radiance are shown in Fig.~\ref{fig:pipeline}.
\secprevspace
\subsection{Unsupervised Pose Estimation}
\label{sec:camera_predictor}
\secpostvspace
As described in \secref{sec:voxel_generator}, we assume that an object movement can be factorized into a set of rigid movements of each individual object's part $p$.
However, detecting 3D part poses, especially in an \emph{unsupervised} way, is a difficult task.
MRAA~\cite{MRAA} shows that estimating 2D parts and their poses in an unsupervised fashion is an under-constrained problem, which requires specialized inductive biases to guide the pose estimation towards the proper solution.
We incorporate such an inductive bias by framing pose prediction as a 2D landmark detection problem which CNNs can solve proficiently due to their natural ability to detect local patterns~\cite{jakab2018unsupervised}.

To lift this 2D bias into 3D, we estimate the poses of 3D parts by learning a set of 3D keypoints in the canonical space and detecting their 2D projections in the current frame using a 2D CNN.
We then use a differentiable  Perspective-n-Point (PnP) formulation to recover the pose of each part since we know its corresponding 2D and 3D keypoints.
More formally, PnP is a problem where, given a set of the 3D keypoints $\kpthree \in \R^{\nk \times 3}$, a set of corresponding 2D projections $\kptwo \in \R^{\nk \times 2}$ and the camera intrinsics parameters, one need to find a camera pose $\T = [ \rot, \tr ] \in \R^{3 \times 4}$, such that $\kpthree$ project to $\kptwo$ when viewed from this pose. 
Note that, while $\T$ represents the pose of the camera with respect to the part, in our framework we consider the camera extrinsics to be constant and equal to the identity matrix, i.e. a part moves while the camera remains fixed.
Recovering a part's pose with respect to the camera is performed by inverting the estimated pose matrix $\Tp = [\rotp, \trp] = [ \rot ^ {-1}, -\rot ^ {-1}\tr]$.

To implement this idea, we introduce $\nk$ learnable canonical 3D keypoints $\kpthree_p$ for each part, totaling $\nk \times \np$.
These 3D keypoints are shared among all the objects in a dataset, which are directly optimized with the rest of the model's parameters. 
Then, we define a 2D keypoints prediction network $\C$, which takes frame $\f_i$ as input and outputs $\nk$ 2D keypoints $\kptwo_p$ for each part $p$, where each 2D keypoint corresponds to its respective 3D keypoint.
The pose of part $p$ is thus recovered as:
\begin{equation}
\label{eq:pnp}
\Tp^{-1} = \text{PnP}\left(\kptwo_p, \kpthree_p\right) = \text{PnP}\left(\C(\f_i), \kpthree_p\right).
\end{equation}
Crucially, in this formulation $\kpthree_p$ are shared for all the objects in the dataset, thus all objects will share the same canonical space for poses.
This property is essential for performing cross-subject animations, where poses are estimated on frames depicting a different identity.

We used the EPnP~\cite{EPnP} implementation from Pytorch3D~\cite{pytorch3d}, since we found it to be significantly faster and more stable than the methods based on declarative layers~\cite{gould2022deep, BPnP}.

\secprevspace
\subsection{Volumetric Skinning}
\label{sec:deformation_modeling}
\secpostvspace

In this section, we describe the procedure to deform the canonical volumetric object representation into its representation in the driving pose.
The deformation can be completely described by establishing correspondences between each point $x_d$ in the deformed space and points $x_c$ in the canonical space.
We establish such correspondence through Linear Blend Skinning (LBS)~\cite{lewis2000pose}:
\begin{equation}
\label{eq:lbs}
x_d = \sum_{p=1}^{\np} w_p^c(x_c)\left(\rot_p  x_c + \tr_p\right),
\end{equation}
where $w_p^c(x)$ is a weight assigned to each part $p$. Intuitively, LBS weights segment the object into different parts.
As an example, a point with LBS weight equal to 1.0 for the left hand, will always move according to the transformation for the left hand.
Unfortunately, during volumetric rendering we typically need to query canonical points using points in the deformed space, requiring solving Eq.~\eqref{eq:lbs} for $x_c$.
This procedure is prohibitively expensive~\cite{li2022tava}, so we rely on the approximate solution introduced in HumanNeRF~\cite{HumanNerf}, which defines \emph{inverse} LBS weights $w_p^d$ such that:
\begin{equation}
\label{eq:lbs_hnerf}
x_c = \sum_{p=1}^{\np} w_p(x_d)\left(\rot_p^{-1}  x_d - \rot_p^{-1}\tr_p\right),
\end{equation}
where weights $w_p^d$ are defined as follows:
\begin{equation}
\label{eq:invlbs}
    w_p(x_d) = \frac{w_p^c(\rot_p^{-1} x_d - \rot_p^{-1} \tr_p)}{\sum_{p=1}^{\np} w_p^c(\rot_p^{-1} x_d - \rot_p^{-1} \tr_p)}.
\end{equation}
This approximation has an intuitive explanation, i.e. given the deformed point, we project it using the inverse $\Tp$ to the canonical pose and check if it corresponds to the part $p$ in canonical pose.
It is easy to see that if each point has a strict assignment to a \emph{single} part and there is no self-penetration in the deformed space, the approximation is exact. In our work, we parametrize $w_p^c$ as the channel-wise softmax of $\VLBS$. Examples of the parts are given in Figs.~\ref{fig:teaser} \&~\ref{fig:pipeline}.

\secprevspace
\subsection{Volumetric Rendering}
\label{sec:volumetric_rendering}
\secpostvspace

We render the deformed object using differentiable volumetric rendering~\cite{NeRF}. Given camera intrinsics and extrinsics, we cast a ray $r$ through each pixel in the image plane and compute the color $c$ associated to each ray by integration as:
\begin{equation}
    c(r) = \int_{t_n}^{t_f} e^{-\int_{t_n}^{t} \sigma(r(s)) ds}\sigma(r(t))c(r(t)) dt, \label{eq:nerf}
\end{equation}
where $\sigma$ and $c$ are functions mapping each 3D point along each ray $r(t)$ to the respective volume density and radiance.
In our framework, we parametrize $\sigma$ as $\Vdens$ and $c$ as $\Vfeat$ which can be efficiently queried using trilinear interpolation.
We train the model using a camera with fixed extrinsics initialized to the identity matrix, and fixed intrinsics.
Note that, to reduce computational resources, we render images directly from voxels without any additional MLP, nor did we employ any upsampling technique.


We assume that the background is flat and it is not moving. We model it as a plate of fixed, high density. This density is modeled with a single dedicated volume, while the color is obtained from $\Vfeat$.

\subsection{Training}
\label{sec:training}
Learning a 3D representation of an articulated object from 2D observations without additional supervision is a highly ambiguous task, prone to spurious solutions with poor underlying geometry that leads to corrupted renderings if the camera is moved away from the origin. 
We devise a two-stage training strategy that promotes learning of correct 3D representations. 
First, we train the model with only a single part, i.e. $\np = 1$. This allows the model to obtain meaningful estimation of the object geometry. Thus we name this pretraining a Geometry phase or \textit{G-phase}.
During the second phase, we introduce $\np = 10$ parts, allowing the model to learn the pose of each part. 
We copy all the weights from the \textit{G-phase}. Moreover, for $\C$ the weight of the final layer is extended such that all the part predictions are the same as in the first stage, while for $\G$, we just add additional weights for $\VLBS$ initialized to zero.


The model is trained using a range of losses.\\
\noindent\textbf{Reconstruction loss.} We use perceptual reconstruction loss~\cite{johnson2016perceptual} as the main driving loss. Similarly to FOMM~\cite{FOMM} we use a pyramid of resolutions:
\begin{equation}
\label{eq:reconstruction}
    \mathcal{L}_{\mathrm{r}} = \sum_{l} \sum_{i} \left|\mathrm{VGG}_i(\mathrm{D}_l \odot \hat{\f}) - \mathrm{VGG}_i(\mathrm{D}_l \odot \f) \right|,
\end{equation}
where $\mathrm{VGG}_i$ is the $i^{\textrm{th}}$-layer of a pretrained VGG-19~\cite{DBLP:journals/corr/SimonyanZ14a} network, and $\mathrm{D}_l$ is a downsampling operator corresponding to the current resolution in the pyramid. The same loss is enforced for $\f^{\mathrm{low}}$.

\noindent\textbf{Unsupervised background loss.} Contrary to 2D frameworks for unsupervised animation that use motion cues to separate background from foreground objects, our generator $\G$ mostly relies on appearance features, thus it is harder for it to disentangle the background from the foreground. In the first stage, we encourage the model to correctly disentangle the background from the foreground leveraging a coarse background mask $B$ that we obtain in an unsupervised manner from MRAA~\cite{MRAA}. Given the occupancy map $O$ for the foreground part obtainable by evaluating Eq.~\eqref{eq:nerf} excluding the background, we enforce a cross entropy loss:
\begin{equation}
    \mathcal{L}_{\mathrm{bkg}} = \sum_i O \log (1 - B) + (1 - O) \log (B),
\end{equation}
the background mask $B$ is very coarse and we observe is necessary only in the earliest iterations to avoid degenerate solutions, thus we reduce the contribution of this loss each epoch, we specify the exact schedule in ~\apref{ap:implementation-details}.

\noindent\textbf{Pose losses.} Finally, to regularize the PnP-based pose prediction we add two regularization terms: equivariance and projection. First one is a standard technique for regularizing unsupervised keypoints~\cite{FOMM,MRAA}:
\begin{equation}
    \mathcal{L}_\mathrm{eq} = \left|A \circ \C(F) - \C(\mathcal{W}(\f, A))\right|,
\end{equation}
where $A$ is a random affine transformation, and $\mathcal{W}$ is a warping operation. The intuition behind this loss is that when an image is deformed, its keypoints should undergo a similar deformation. Second, we explicitly minimize the $\kpthree$ reprojection error with $\kptwo$:
\begin{equation}
    \mathcal{L}_\mathrm{proj} = \sum_p \left|\kptwo_p - \Pi (\kpthree_p, \Tp) \right|,
\end{equation}
where $\Pi_p$ projects the points according to the estimated camera pose $\Tp$. This loss enforces keypoints to comply with the predicted camera transformation $\Tp$, improving the stability of the PnP algorithm.

The final loss is the sum of all terms with equal weights. Note that for the second stage $\mathcal{L}_{\mathrm{bkg}}$ is not used.

\subsection{Inference}
\label{sec:inference}

Despite our model learns the embedding space for the identities in the training set only, it can be used to model previously unseen identities. Given an image of an unseen identity $\f_{\mathrm{test}}$ and a randomly initialized embedding $\emb_{\mathrm{test}}$, we optimize the reconstruction loss $\mathcal{L}_{\mathrm{r}}$ (Eq.~\ref{eq:reconstruction}) with respect to the embedding $\emb_{\mathrm{test}}$. This procedure produces a volumetric representation with detailed geometry, but imperfect textures. We address this issue by finetuning the generator $\G$, following the pivotal tuning procedure~\cite{PTI}. In order to avoid significant distortions to the geometry, during this finetuning stage we regularize $\Vdens$ and  $\VLBS$ to stay close to their values prior to finetuning. Note that we only optimize with respect to the appearance and do not modify the 2D keypoint predictor $\C$, ensuring that motion can be transferred from different objects. Additional details concerning this embedding are provided in \apref{ap:implementation-details}.

\secprevspace
\section{Experiments}
\secpostvspace

Evaluating animation, whether 2D or 3D, is a challenging task as there is no ground truth for the animated images. We are not aware of prior works in unsupervised volumetric animation, hence, in this section we establish an evaluation protocol for this task. Our protocol makes use of established metrics in unsupervised 2D animation, when applicable, and introduces procedures to evaluate the quality of the synthesized 3D geometry and animation under novel views. 

\begin{figure*}
\centering
\begin{overpic}[scale=0.5,unit=1mm,width=\linewidth]{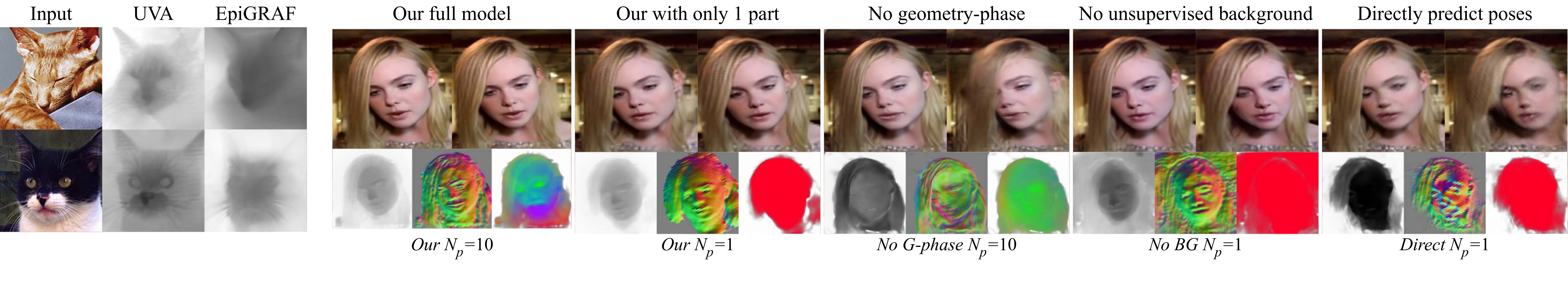}
\put(2.5, 0){\small(a) Depth comparisons}
\put(73, 0){\small(b) Qualitative ablation results of methods in Tab.~\ref{tab:ablation_results}}
\end{overpic}
\vspace{-0.5cm}
\caption{(a) Typical depth examples of embedded images using our method (UVA) and EpiGRAF~\cite{EpiGRAF}. Note, UVA’s depth contains sharper details regardless of the pose. (b) We show a block for each method, with novel views (top), and depth, normals, parts (bottom).
}
\vspace{-0.2cm}
\label{fig:cats-and-ablations}
\end{figure*}

\noindent{\textbf{Datasets}}. To evaluate our method we use three publicly available datasets: 1) Cats~\cite{Cats_dataset}, consisting of 9,993 images of cat faces. We used 9,793 for training and 200 for testing. 2) For VoxCeleb~\cite{voxceleb}, we employed the same pre-processing as FOMM~\cite{FOMM}, using 19522 face videos for training and 100 for testing. 3) TEDXPeople~\cite{stylepeople} is a video dataset of TEDx speakers. Using timestamps provided by~\cite{stylepeople}, we extract continuous video chunks. More details can be found in \apref{ap:dataset-details}.
In total, we employ 40896 videos for training, and retain 100 videos for testing.
\begin{figure*}
\centering
\includegraphics[width=\textwidth]{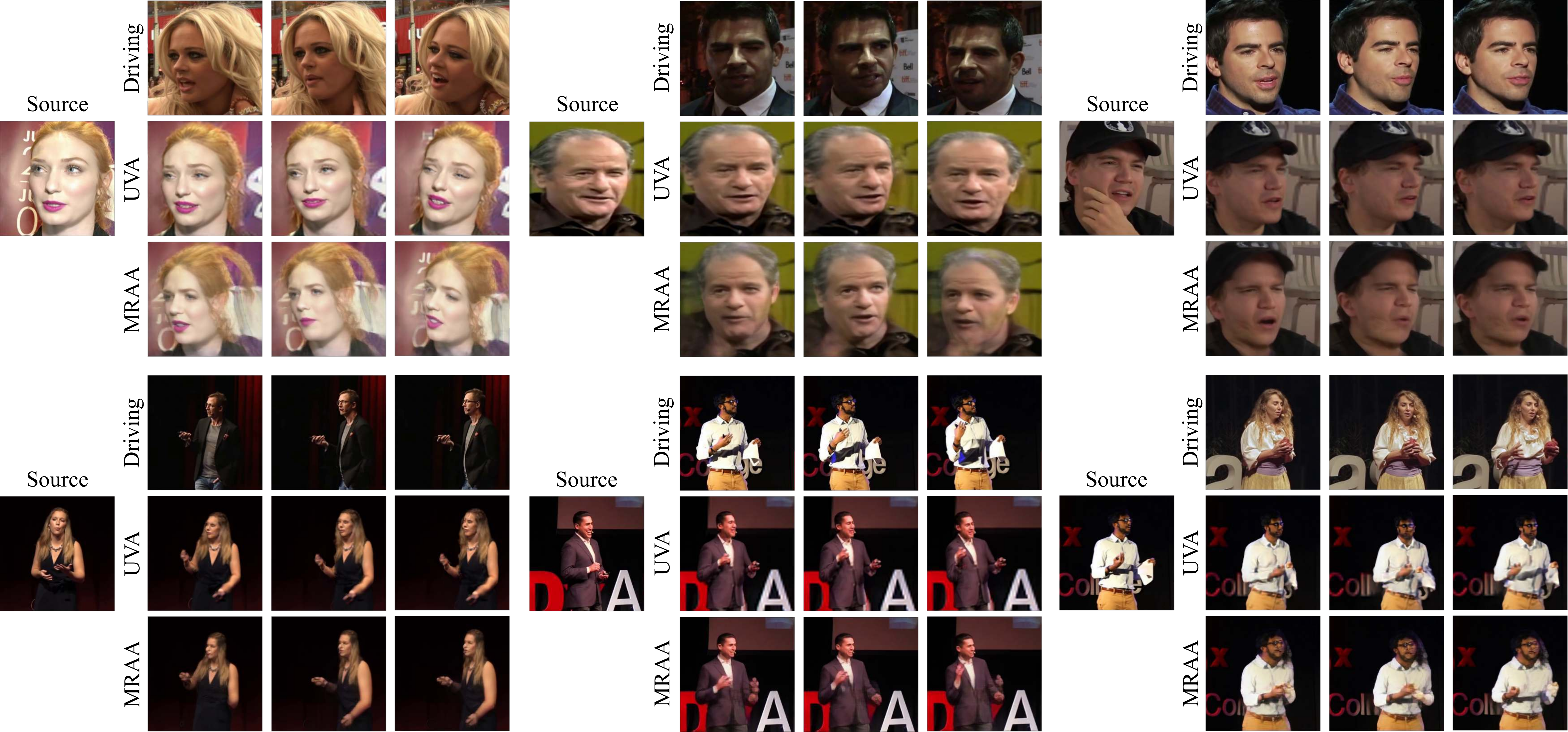}
\vspace{-0.6cm}
\caption{\textbf{2D animation.} Example 2D animations on bodies and faces from our method (UVA) and a state-of-the-art work, MRAA~\cite{MRAA}. As UVA models objects in canonical 3D space, it better preserves an object's shapes when animated. Zoom-in for greater detail.
}
\vspace{-0.5cm}
\label{fig:mraa-comparisons}
\end{figure*}

\secprevspace
\subsection{Geometry from Image Data}
\secpostvspace
Our method learns high-fidelity geometry from images or videos \emph{without camera or geometry} supervision. This is a challenging setting, even for recent 3D-GANs, as they \emph{require} camera supervision. In this setting, we compare the quality of inferred geometry to a state-of-the-art 3D-GAN, EpiGRAF~\cite{EpiGRAF}, trained with ground truth camera poses.
As both UVA and EpiGRAF render non-absolute depth, to evaluate its quality, we use the Pearson correlation coefficient.
Given a test image, we reconstruct it by inversion, and obtain depth using volumetric rendering.
We then measure the correlation between the predicted depth and the depth estimated with an off-the-shell depth estimator~\cite{Omnidata}. For a fair comparison, during inversion, we do not provide camera poses to EpiGRAF and instead find them during the optimization, in combination with the rest of the parameters.
UVA provides higher-quality depth, while not requiring camera supervision during training, reaching a correlation value of $0.63$. EpiGRAF reaches only $0.53$, often failing to render accurate depth for non-frontal cameras (see Fig.~\ref{fig:cats-and-ablations}a).




\begin{figure*}
\centering
\includegraphics[width=\textwidth]{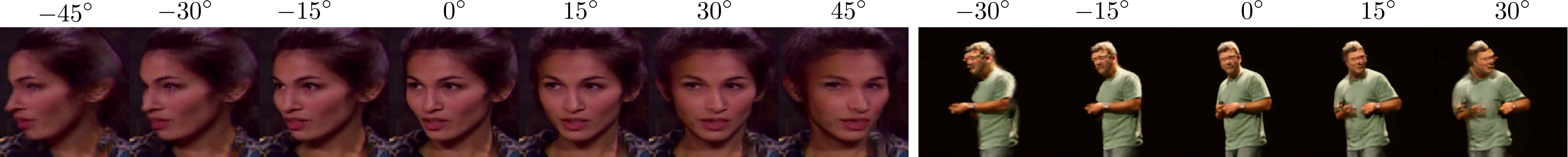}
\vspace{-0.5cm}
\caption{\textbf{Novel view synthesis.} We report typical examples of novel views synthesized using a single input image. For bodies, we show a narrower range, as the TEDXPeople dataset is biased towards frontal poses. 
\vspace{-0.3cm}
}
\label{fig:rotations}
\end{figure*}

\secprevspace
\subsection{Animation Evaluation}
\secpostvspace
\label{sec:animation-eval}
Unsupervised animation in 3D is a new task introduced in this work. A key feature of 3D animation is the ability to change the viewpoint from which the object is rendered during animation. Commonly used animation datasets, however, do not typically offer multi-view data. To evaluate viewpoint consistency without access to multi-view data, we introduce three new metrics: Average Yaw Deviation (AYD), Average Shape Consistency (ASC), and Average Pose Consistency (APC). In more detail, given an object, we rotate it along the y-axis using a set of predefined angles. We then fit an SMPL~\cite{choi20223dcrowdnet} model for humans and a 3DMM~\cite{feng2021deca} for faces to the frontal and rotated views of the objects. These models estimate the root angle, defining how the object is oriented with respect to the camera; a shape parameter, defining the identity of the object; and a parameter defining its pose (in terms of joint rotations for SMPL and facial expression parameters for 3DMM). To evaluate the ability of the model to rotate the object by the required angle to produce novel views, we use AYD. In particular, we compute the y-axis component of the root angle between the rotated and non-rotated object, and compare it with the known yaw of the camera, used to render that view. We use ASC to compare the consistency of the shape parameters between the frontal and the rotated views. A lower ASC indicates that the identity is better preserved during rotation. APC is used to measure how much the pose is altered during rotation, with a lower APC indicating better preservation of the object pose. These metrics enable evaluating the capabilities of competing models in generating view-consistent results. \apref{ap:metric-details} contains full details on these metrics.

\begin{table*}[t]
    \setlength{\tabcolsep}{6.0pt}
    \footnotesize
    \centering
    \begin{tabular}{c|cccccc|cccccc}
        \toprule
        &\multicolumn{6}{c}{VoxCeleb} & \multicolumn{6}{c}{TEDXPeople} \\
        \midrule
        Method & AYD$\downarrow$ & ASC$\downarrow$ & APC$\downarrow$ & L1$\downarrow$ & AKD$\downarrow$ & AED$\downarrow$ & AYD$\downarrow$ & ASC$\downarrow$ & APC$\downarrow$  & L1$\downarrow$ & (AKD$\downarrow$, MKR$\downarrow$) & AED$\downarrow$ \\
        \midrule
         FOMM \cite{FOMM} & 0.655 & 0.129 & 0.177 & \textbf{0.0413} & 1.289 & 0.134 & 0.507 & 0.028 & 1.07 & 0.0318 & (3.248, 0.009)  & 0.120   \\
         MRAA \cite{MRAA} & 0.173 & 0.123 & 0.174 & 0.0424 & \textbf{1.250} & 0.131 & 0.181 & 0.023 & 0.702 & \textbf{0.0262} & (\textbf{2.282}, \textbf{0.007}) & 0.101 \\
         LIA \cite{LatentImageAnimator}  & 0.207 & 0.130 & 0.190 & 0.0529 & 1.437 & 0.138 & - & - & - & - & - & - \\
         Our 1 frame & 0.051 & \textbf{0.078} & 0.144 & 0.0655 & 1.737 & 0.226 & 0.128 & \textbf{0.019} & 0.635 & 0.0474 & (3.571, 0.017) & 0.163 \\
         Our 5 frame & \textbf{0.045} & 0.091 & \textbf{0.112} & 0.0418 & 1.378 & \textbf{0.111} & \textbf{0.107} & 0.021 & \textbf{0.571} & 0.029 & (2.373, 0.014) & \textbf{0.086} \\
         \bottomrule
    \end{tabular}
    \vspace{-0.3cm}
    \caption{Comparison with 2D animation methods. Novel view synthesis for AYD, ASC \& APC from yaw in range $-45^{\circ}$ to $+45^{\circ}$. }
    \label{tab:animation2d}
    \vspace{-0.3cm}
\end{table*}
No prior unsupervised animation method~\cite{FOMM,MRAA,LatentImageAnimator} offers a built-in ability to generate the data under novel views. Thus, for~\cite{FOMM,MRAA} we introduce a simple, depth-based method to generate novel views. First, we predict the depth from a monocular depth predictor~\cite{Omnidata} and normalize it to make it compatible with our camera intrinsics. Then, for each method, we estimate parts and their affine transformations. We choose a central 2D keypoint for each part, and augment it with 4 additional keypoints in its neighborhood. Using the depth, we lift the keypoints in 3D and re-project them into the novel viewpoint.
From these new keypoints, a new affine transformation is estimated and used to drive the view synthesis.
We then evaluate against LIA~\cite{LatentImageAnimator}, which expresses animation as linear navigation in a latent space.
Interestingly, for the VoxCeleb~\cite{voxceleb} dataset, we found one of the components of its latent space to correlate with the rotation of the head along the y-axis. Exploiting this finding, we fit a linear model mapping the magnitude of the movement along this latent component to the produced head rotation, and use it to generate the head under novel viewpoints. 

We also use the standard 2D reconstruction metrics: L1, AKD/MKR~\cite{MonkeyNet}, AED~\cite{MonkeyNet}.
However, we emphasize that such metrics favor 2D methods, which can solve the 2D animation problem by copying pixels from the source to the target view, at the cost of limited 3D understanding and consistency. In contrast, UVA renders view-consistent pixel values from a 3D representation, making this shortcut unavailable.
A significant gap may also be introduced by the single-image embedding procedure we adopt.
Note, however, that as our embedding procedure seamlessly supports the use of multiple source frames at inference time, a shared representation can be optimized, pooling information from all available frames to improve performance.
We demonstrate the results of our model with one and five frames. We also note that, despite a wide range of a viewpoints in different videos, subjects in each individual video in VoxCeleb and TEDXPeople have very limited 3D rotations, as they primarily face towards the camera.
Thus, standard reconstruction metrics do not reflect the model's capacity to perform complex 3D transformations.

We provide the quantitative results in Tab.~\ref{tab:animation2d}. As the affine transformations of FOMM~\cite{FOMM} are mostly based on edge detection that is not very robust, any minor modification of this transformations for novel view synthesis leads to significant movement.
Thus, FOMM has the worst AYD among all methods.
Affine estimation in MRAA, in contrast, is significantly more robust, and thus it has a significantly lower AYD.
However, we observe that MRAA does not have enough knowledge about the 3D structure of the objects, treating them 
as planes---while they roughly preserve the shape and pose for small angles, for larger angles objects become thinner, until 
they eventually disappear.
LIA has a rotation direction that is entangled with the other movements, and thus it has the lowest ASC and APC.
Finally, our model is the best at preserving shape and expressions, as judged by the ASC and APC. Moreover, our model also provides the most meaningful rotations as judged by the AYD.
When standard reconstruction metrics are considered, our 5 frame model performs on par with the baselines.
However, as previously mentioned, these metrics do not reflect the ability of the model to preform complex 3D movements.
This point is further highlighted in Fig.~\ref{fig:mraa-comparisons}, when the pose of the source and driving images differ significantly, MRAA fails to properly reflect that, while our model produces more consistent results.
Interesting, we also note that, as our model is based on learning a 3D prior and not copying pixels, it can filter out some occlusions, as seen in the third column in Fig.~\ref{fig:mraa-comparisons}, while MRAA produces artifacts in the occluded region.






\subsection{Ablation Studies}
\begin{table}[t]
    \setlength{\tabcolsep}{5.0pt}
    \footnotesize
    \centering
    \resizebox{\linewidth}{!}{%
    \begin{tabular}{c|cccccc}
        \toprule 
        Method & AYD$\downarrow$ & ASC$\downarrow$ & APC$\downarrow$ & L1$\downarrow$ & AKD$\downarrow$ & AED$\downarrow$ \\
        \midrule
         \textit{Direct} $\np = 1$ & 0.707 & 0.160 & 0.239 & 0.0723 & 3.582 & 0.326 \\
         \textit{No BG} $\np = 1$ & 0.301 & 0.117 & 0.216 & 0.0702 & 2.410 & 0.263\\
         \textit{Our} $\np = 1$ & 0.141 & 0.113 & 0.210 & 0.0637 & 2.170 & 0.242 \\
         \midrule
         \textit{No G-phase} $\np = 10$ & 1.08 & 0.145 & 0.226 & \textbf{0.0620} & 1.993 & 0.243 \\
         \textit{Our} $\np = 10$ & \textbf{0.051} & \textbf{0.078} & \textbf{0.144} & 0.0655 & \textbf{1.737} & \textbf{0.226} \\
         \bottomrule
    \end{tabular}
    }
    \vspace{-0.3cm}
    \caption{Ablation results on the VoxCeleb dataset.}
    \label{tab:ablation_results}
    \vspace{-0.3cm}
\end{table}
We evaluate the key design choices made in our framework. First, we compare our PnP-based part pose predictor with direct part pose prediction (\textit{Direct}). As directly predicting an $\R^{3\times3}$ rotation matrix could produce solutions not corresponding to a rigid rotation, we adopt the 6D rotation parameterization from~\cite{zhou2019rotations}. The geometry learned by this approach is essentially flat. We compare our method and \textit{Direct} only in the geometry phase of training (e.g., when $\np = 1$), as it does not produce sufficiently accurate geometry to proceed with the next phase. We also demonstrate the effect of the unsupervised background loss $\mathcal{L}_\mathrm{bkg}$ by training the model without this loss (\textit{No BG}). Finally, we investigate the importance of two-phase training, learning a model with multiple parts without the geometry phase \textit{No G-phase}. We show numerical results in Tab.~\ref{tab:ablation_results} and qualitative examples in Fig.~\ref{fig:cats-and-ablations}b. Our full model achieves the best scores, and generates higher fidelity novel views and geometric details.
The utility of the geometry phase is clearly demonstrated by the scores and qualitative results, which, without this phase, produce corrupted results and do not learn representative parts.
While it produces meaningful depth, the model trained without $\mathcal{L}_\mathrm{bkg}$ fails to separate the background and foreground.

\secprevspace
\subsection{Geometry Evaluation for Synthetic Objects}
\secpostvspace

\begin{figure}[t]
\centering
\begin{tabular}{c}
     \includegraphics[width=0.95\linewidth]{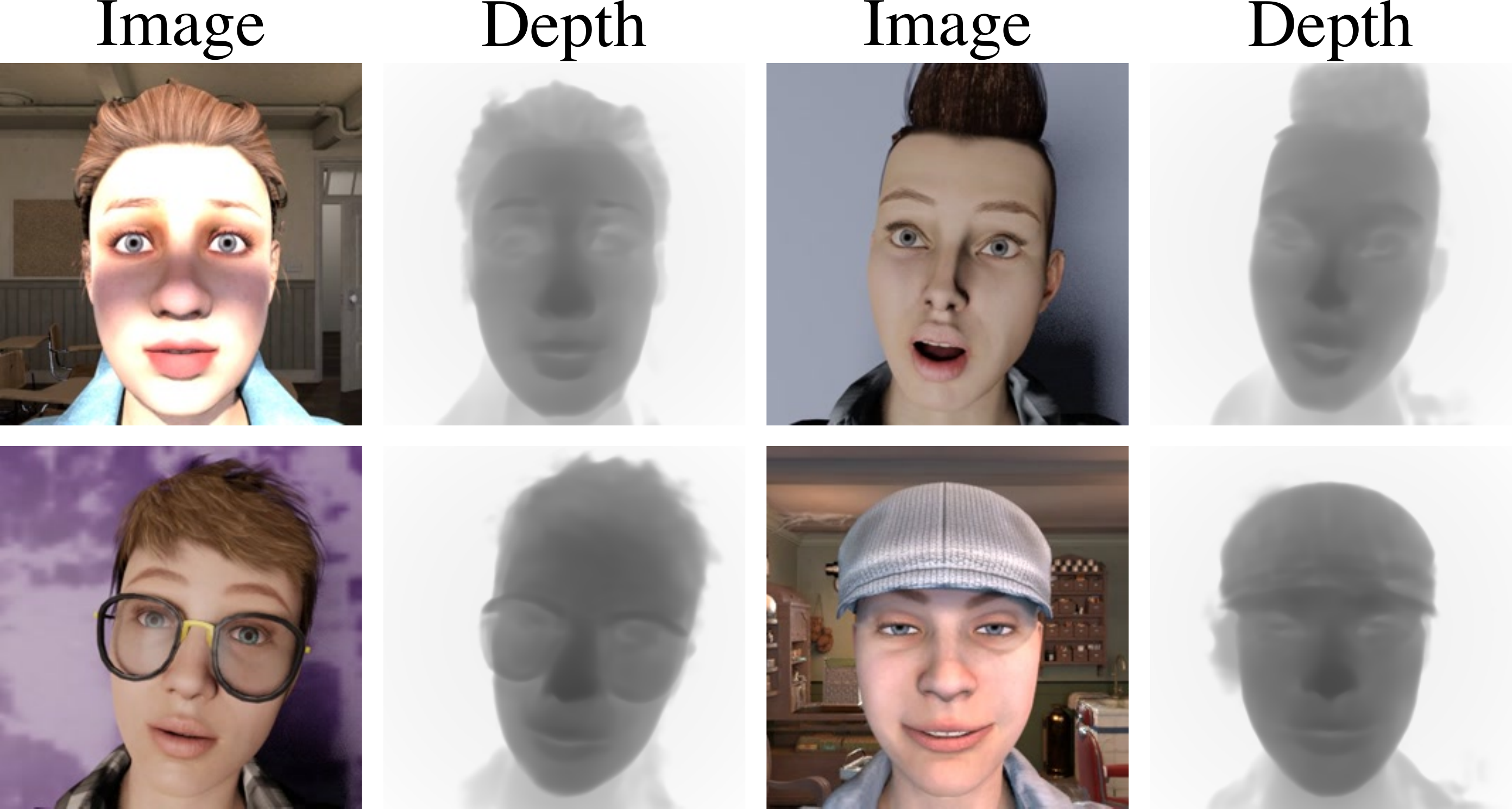} \\
     (a) Khan \etal~\cite{KHAN2021479} \\
     \includegraphics[width=0.95\linewidth]{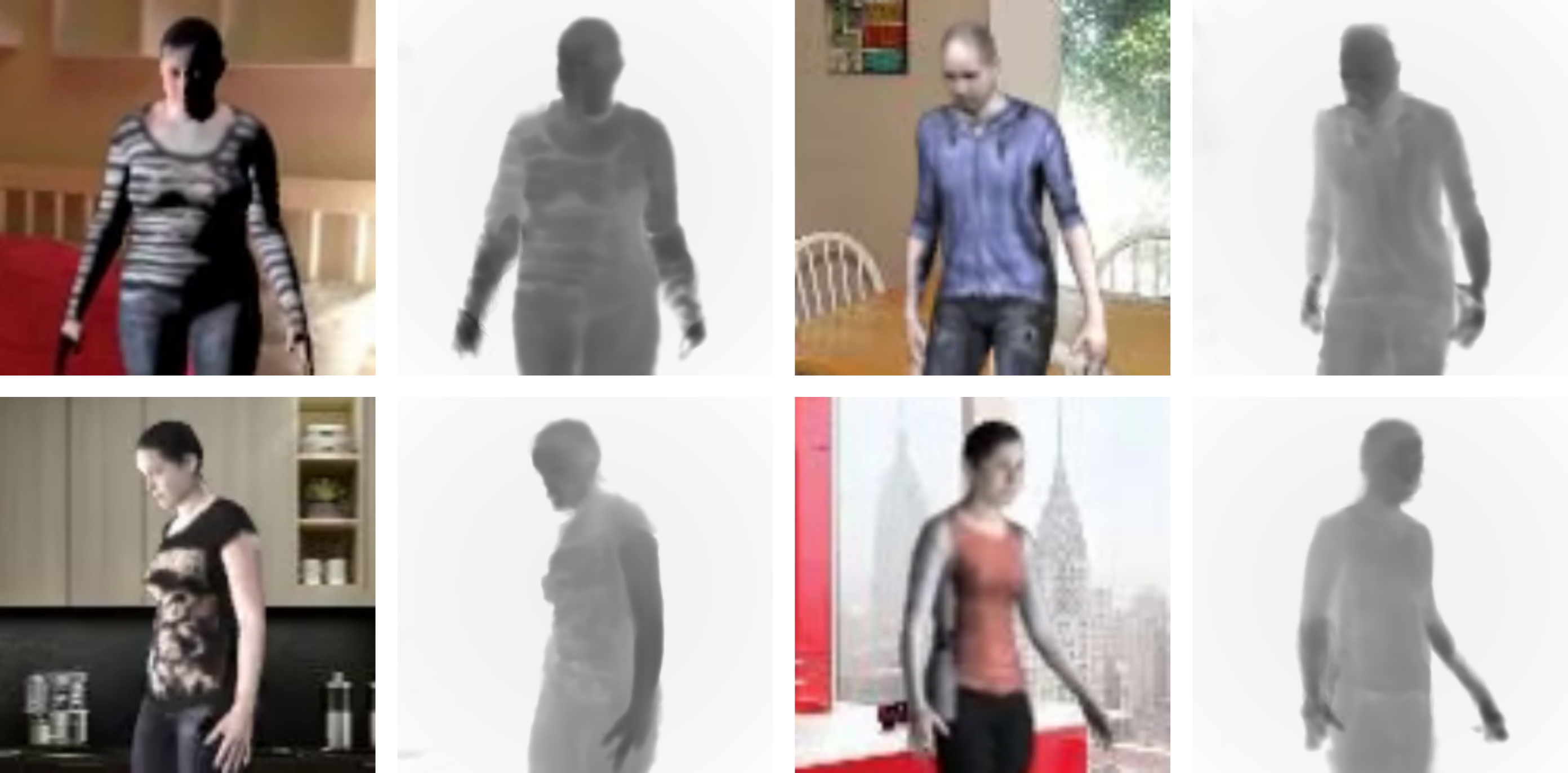} \\
     (b) SURREAL~\cite{varol17_surreal}
\end{tabular}
\caption{Visualization of predicted depth for synthetic datasets. We show the input image and the depth predicted by our method.}
\label{fig:synthetic}
\vspace{-0.3cm}
\end{figure}

To further evaluate the quality of the learned geometry, we ran experiments on images from two synthetically rendered datasets providing ground truth depth: 1.) that of Khan~\etal~\cite{KHAN2021479}, which provides high-quality, portrait-style facial images; and 2.) SURREAL~\cite{varol17_surreal}, which provides full-body renderings of animated subjects. We use 112 image for faces and 60 images for bodies, cropped such that they roughly correspond to the cropping used in the respective real datasets used for training. 
These datasets contain subjects with widely varying identities, poses, hairstyles, and attire, rendered in different environments and lighting conditions.
However, for these experiments we did not rely on synthetic data for training, instead using models pretrained on 2D images from VoxCeleb~\cite{voxceleb} or TEDXPeople~\cite{stylepeople} for faces or bodies, respectively.
Despite the domain gap between our training and evaluation data, we are able to obtain high-quality depth estimates for these synthetic renderings using models trained only on real, in-the-wild images.
Given a synthetic input image, we invert it, then compute the Pearson correlation coefficient between our method's inferred depth and the ground truth. 
For these experiments, as we are only concerned with the geometry of the target object, we masked out the depth for background regions, computing the correlation only between the depths of foreground pixels.
We compare our predicted depth with the general purpose state-of-the-art depth predictor Omnidata~\cite{Omnidata}.
The depth correlation for Omnidata is 0.602 for faces and 0.470 for bodies, while for our method they are 0.793 and 0.568, respectively. In Fig.~\ref{fig:synthetic}, we show the image along with the reconstructed depth. 
These results demonstrate that our unsupervised method learns meaningful geometric representations, even for significantly out-of-distribution inference data.

\secprevspace
\section{Limitations}
\secpostvspace

Our model addresses, for the first time, the task of unsupervised 3D animation. While our model obtains compelling results on this challenging task, here we note some limitations:

\begin{itemize}
    \item Our method assumes the object can be represented with a voxel cube of size $64^3$. We notice that when generating novel views involving large camera displacements from the original pose, some seam-like artifacts may appear. We believe they are due to the small size of the voxel cube and errors in predicting precisely the distance of the part, which could lead to a slight displacement between different parts.
    \item For each test identity, our model makes use of an optimization-based procedure to compute the respective identity embedding and fine-tune the generator. This procedure increases the inference cost of our model, but needs to be performed only once for each test identity, thus the cost of the procedure is amortized when producing a large number of frames.
    \item Our model renders frames at a resolution of $256\times256$, which is lower than the ones typically supported by state of the art 2D animation methods. This is a common limitation of 3D methods based on volumetric rendering, and we expect continuous progress in efficient volumetric representations and rendering to enable the generation of higher resolution images.
    \item Our method can learn geometry only from the views that were observed in the training dataset, thus, for the back side of the face in VoxCeleb~\cite{voxceleb} and the back side of the body in TEDXPeople~\cite{stylepeople}, no precise geometry is learned.
\end{itemize}

\secprevspace
\section{Conclusion}
\secpostvspace
Our approach for unsupervised volumetric animation demonstrates a significant step towards 3D animation of dynamic objects. 
While trained exclusively on real-world monocular 2D videos, our method obtains high quality geometry, object parts, 3D segmentation and normals. Due to the unsupervised nature of our work, the same approach applies to a variety of object categories without using explicit labels or other cumbersome supervision. 
This understanding of the underlying geometry and structure of the object, allows our method to perform animation and novel view synthesis at the same time. 
These properties open exciting possibilities for employing this information for future exploration, \eg controlling an object's fine-grained shape, or relighting it for composition into novel environments.

\clearpage

{\small
\bibliographystyle{ieee_fullname}
\bibliography{ref}
}
\clearpage

\appendix


\secprevspace
\section{Implementation details}\label{ap:implementation-details}
\apppostvspace

\textbf{Network architectures.} Similar to FOMM~\cite{FOMM}, for our keypoint predictor $\C$ we employ a U-Net~\cite{unet} backbone operating on $64 \times 64$ resolution input images.
Following FOMM~\cite{FOMM}, the architecture consists of five "convolution - batch norm - ReLU - pooling" blocks in the encoder and five "upsample - convolution - batch norm - ReLU'' blocks in the decoder.
For the generator $\G$, we use a simple decoder architecture. We use embedding size $\nemb=64$ for all the datasets. In order to regularize the embeddings, we apply differentiable whitening~\cite{WCGAN}, which we found to provide  additional training stability.  Given the embedding $\emb$, we transform it to a $4^3$ voxel cube with $512$ features using a fully connected layer. This voxel cube is then passed through four 3D-Residual Blocks with upsampling. The 3D-Residual block consists of two $3 \times 3 \times 3$ convolutions and two batch normalization in the main branch, and one $1 \times 1 \times 1$ convolution in the residual branch. We choose ReLU for our non-linearity. Each 3D-Residual block reduces the number of features two times, while also increasing the resolution two times using nearest neighbor upsampling. After the final upsampling layer, the voxel cube has size $64^3$ and $32$ features. We make use of a final projection layer implemented as a $1\times1\times1$ convolution preceded by batch normalization to obtain $1 + 3 + 10$ features for density, RGB, and LBS volumes, respectively. 

\textbf{Neural rendering.} For neural rendering, we use fixed camera extrinsics equal to the identity matrix, and a fixed intrinsics matrix assuming a field of view of $0.175$~\cite{Giraffe}. Based on this configuration, we define the region in which we sample points for rendering to be the cube spanning the region $[-1.0088, 1.0088] \times [-1.0088, 1.0088] \times [9.5000, 11.5000]$. The camera in our settings looks in the positive z direction. We call this region the rendering cube. We use the rendering cube intersections with casted rays to define the near and far camera planes, and uniformly sample $128$ points between them. We note that during the $G-phase$, there is no need to capture small parts, as a single part is employed. Thus, to speed up training, we reduce the number of sampling points to $48$ in this phase. When mapping the rendering cube to the respective volumes, we increase the rendering cube by a factor of 1.075. For the Cats~\cite{Cats_dataset} dataset, we scale the rendering cube by a factor of 1.2. In this manner, we increase the amount of actual space that can be covered by the modeled volume, allowing for a larger set of transformations. Following NeRF~\cite{NeRF}, we apply perturbations to the sampled point in two ways during training. First, we perturb the position of each point along the ray. Second, we add noise to the sampled densities before rendering. We use a standard deviation of $0.5$ for the noise. As the geometry is mostly discovered at the begining of training, we linearly decrease it during each phase of training down to zero at $100k$ training steps.

\textbf{Perspective-n-Point.} We obtain a solution to the PnP problem leveraging the differentiable EPnP~\cite{EPnP} implementation from PyTorch3D~\cite{pytorch3d}. For each part $p$, we predict $\nk = 5^3 =125$ keypoints, for a total of $\np \times \nk = 125 \times 10 = 1250$ keypoints. Each keypoint is represented with an unconstrained three-dimensional parameter, followed by sigmoid and normalization to ensure that each individual keypoint always lies inside the rendering cube. We initialize 3D keypoints for each part such that they form a regular cubical grid with $5$ equally spaced keypoints on each side of the cube. Due to possible incoherent arrangements between 2D and 3D keypoints, the estimated part poses may correspond to object positions outside the rendering box. This behavior would prevent the affected parts from learning density, since they would bring no contribution to the rendered image. To address this issue, we add an additional loss that pushes the estimated pose of parts with no associated density to the the pose of the largest part (i.e. with the most density). We formulate this loss as follows:
\begin{equation*}
    \mathcal{L}_{\mathrm{init}} = \sum_p max(0, t - \sigma_p) \vert \Tp - T_{p_{max}} \vert, 
\end{equation*}
where $t=0.01$ is a density threshold, $\sigma_p$ is the density associated with part $p$ (which is the mean of all densities for all sampled points, multiplied by the LBS weight) and $T_{p_{max}}$ is the pose of the part with the maximal density. For the first phase, when a single part is learned, we use the identity matrix in place of $T_{p_{max}}$.

\textbf{Training.} In the first phase, we train the model for 200 epochs with a batch size of 128 on 8 A100 GPUs. To equally balance the contribution of each identity, each epoch consists of a single frame randomly sampled from each video in the dataset. In this phase, we decrease the contribution of the $\mathcal{L}_{\mathrm{bkg}}$ by a factor of 0.8 every 10 epochs. For the Cats~\cite{Cats_dataset} dataset, the method is trained for 1000 epochs on a single A100 GPU with a batch size of 16. In the second phase, we train the model for 1000 epochs using a batch size of 16 on 8 A100 GPUs. The learning rate is decayed 10 times at 600 epochs and 900 epochs. Finally, to encourage the network to discover small parts such as hands in TEDXPeople~\cite{stylepeople} dataset, during this phase we also employ co-part segmentation obtained without supervision from MRAA. The loss consists in the cross-entropy loss between MRAA~\cite{MRAA} co-parts and our rendered LBS weights. As with $\mathcal{L}_{\mathrm{bkg}}$, this loss is decreased by a factor of 0.8 every 10 epochs. For both phases, we use Adam~\cite{Adam} with $lr=5e-4$ and $\beta=(0.5, 0.999)$.

\textbf{Inference.} To embed test images, we rely on a two stage procedure, similar to PTI~\cite{PTI}. First, we optimize the embedding $\emb$ for a particular image using the reconstruction loss $\mathcal{L}_\mathrm{r}$ used during training, and employing Adam~\cite{Adam} with $lr=1e-2$. The optimization is run for 3000 steps, and the learning rate is decayed by a factor of 10 every 750 steps. Similarly to StyleGAN2~\cite{StyleGAN2}, we add random noise to the embedding to promote exploration. We select an initial standard deviation of 0.5 for this noise and decay it until reaching zero at step 1500. For the second stage, we fine-tune all the generator parameters. To avoid forgetting the useful geometry prior learned during training, we add an additional geometry regularization loss:
\vspace{-0.05cm}
\begin{equation*}
        \mathcal{L}_{\mathrm{geo}} = \vert \hat{\V}^{\mathrm{Density}} - \Vdens \vert + \vert \hat{\V}^{\mathrm{LBS}} - \VLBS \vert, 
\end{equation*}
\vspace{-0.05cm}
where $\hat{\V}^{\mathrm{Density}}$, $\hat{\V}^{\mathrm{LBS}}$ are the density and LBS weights from the first stage, respectively. We also employ additional data augmentation at alternate steps by applying random Euclidean transformations for the source image, for which we sample rotation angles and translations in the $[-0.1, 0.1]$ range. Note that, as our model assumes the background is static, we only enforce this loss for the foreground using the rough background mask obtained from the first stage. We train for 500 steps in this stage. For optimization with 5 input frames, we increase the number of steps in the second stage to 3000. Since 5 images may not fit into a single GPU, we use a batch size equal to 2. Inverting one image takes roughly 10 minutes on an A100 GPU, while inverting five images takes roughly 20 minutes. Our PnP-based part pose estimation algorithm introduces some instability in the estimation of the distance of the part from the camera. While this instability does not produce artifacts when rendering the object from limited rotation angles, it becomes more noticeable when rendering from extreme camera angles. To mitigate such effects, we devise an inference-time filtering strategy to smooth abrupt changes in the estimated depth of each part. For each part, we estimate the distance from the camera origin to the center of the rendering cube. We then compute the mean of all these distances for each part $l_p$. Finally, we rescale the vector from the camera origin to the center of the rendering cube, such that they have the same length $l_p$ in all frames.

\appprevspace
\section{Dataset details}\label{ap:dataset-details}
\apppostvspace
We employ three training datasets: VoxCeleb~\cite{voxceleb}, TEDXPeople~\cite{stylepeople} and Cats~\cite{Cats_dataset}. We adopt the VoxCeleb~\cite{voxceleb} preprocessing of FOMM~\cite{FOMM}, and preprocess Cats~\cite{Cats_dataset} in the same way as~\cite{GRAM, EpiGRAF}. For TEDXPeople~\cite{stylepeople}, we first download the videos listed in~\cite{stylepeople}, then, using the provided timestamps, select continuous chunks of videos starting at the provided timestamp and lasting at most 512 frames. In each chunk, we detect human keypoints and bounding boxes for each frame using~\cite{detectron2}. We clamp the predicted bounding box at the hip joints at the bottom of the frame, then increase its size by a factor of 1.2 so as to capture the subject's full upper body, then make it square. We process the video chunk frame-by-frame, adding each processed frame to the current video sample. If, in some frames, the human is not detected or the bounding box moves significantly from the initial position, we stop the current video sample and start collecting a new sample at the next detection of a human. To further clean the dataset, we discard video samples that are too short (less than 64 frames), to small (less than 256 pixels on any side), have significant background movement (detected using simple $\mathcal{L}_1$ error on pixel values), have no movement in foreground (which most likely indicate that the detected human is a static image visualized during the presentation) or have a width similar to the height (which indicates a failure of the hip predictor). We select only views marked as "front" in the original annotations~\cite{stylepeople}, and from each YouTube video, we take at most three different samples. Our final dataset consists of 40896 different samples from 17451 different YouTube videos. 

\appprevspace
\section{Metric details}
\label{ap:metric-details}
\apppostvspace

A critical aspect of 3D animation is the ability to synthesize novel views of the observed target object. However, evaluating this ability is challenging, as animation datasets typically lack multi-view observations. We thus introduce metrics that, given a triplet composed of a source frame, a driving frame, and a result rendered under a target camera, can quantitatively evaluate the quality of the rendered novel view:
\begin{itemize}
    \item Average Yaw Deviation (AYD): this evaluates whether the object is rendered from the target camera perspective. Given the yaw angle between the camera and the object in the driving frame $\yaw_d$ and in the rendered frame $\yaw_r$, and the yaw angle of the novel view camera with respect to the original camera $\yaw_c$, we define $\mathrm{AYD} = | \yaw_d - (\yaw_c + \yaw_r) |$
    \item Average Shape Consistency (ASC): this evaluates whether the identity of the rendered object is the one in the source frame. Given an identity code for the source frame $\shapecode_s \in \mathcal{R}^{\shapecodelen}$ and an identity code for the rendered frame $\shapecode_r$, we define $\mathrm{ASC} = \frac{|\shapecode_s - \shapecode_r|}{\shapecodelen}$
    \item Average Pose Consistency (APC): this evaluates whether the object is rendered in the pose given by the driving frame. Given a pose code for the driving frame $\posecode_d$ and a pose code for the rendered frame $\posecode_r \in \mathcal{R}^{\posecodelen}$, we define $\mathrm{APC} = \frac{|\posecode_d - \posecode_r|}{\posecodelen}$
\end{itemize}
We obtain $\yaw$, $\shapecode$ and $\posecode$ in a way that is specific to the given object category.
For faces, we compute the head yaw angle $\Theta$ using the 6DOF head pose estimator 6DRepNet \cite{hempel20226d}, which we find robust to extreme head poses and possible corrupted regions in the rendered frames. Given an image, the model directly provides the estimated yaw angle, which we convert to radians prior to the computation. To compute $\shapecode$ and $\posecode$ we use the DECA \cite{feng2021deca} 3DMM. We select this model due to its robustness to large head rotations, its fast, encoder-based inference, and its ability to disentangle the head shape from the current expression. In particular, we define $\shapecode$ as the inferred $\shapecodelen=100$ FLAME~\cite{li2017flame} face shape parameter, which encodes the identity of the subject. We define $\posecode$ as the concatenation of the inferred 50 expression parameters with the estimated jaw rotation in axis-angle representation for $\posecodelen=53$, which together capture the particular facial pose. We choose not to make use of the estimated head yaw angle, since we find it less robust than the one inferred from our adopted 6DOF head pose estimator.
For human bodies, we fit the SMPL~\cite{loper2015smpl} body model to each frame using 3DCrowdNet~\cite{choi20223dcrowdnet}. We choose 3DCrowdNet due to its fast inference time and its robustness to partially-occluded subjects which are frequent in the TEDXPeople~\cite{stylepeople} dataset, where only the upper half of the body is typically present in the frame. 3DCrowdNet requires a set of 2D human body keypoints to be detected for each frame. We first detect person bounding boxes using Faster R-CNN~\cite{ren2015faster} and use VitPose~\cite{xu2022vitpose} to detect the 2D human body keypoints, which we find to work robustly even in the presence of artifacts in the images. Given the fitted SMPL model, we define $\shapecode$ as the inferred $\shapecodelen=10$ body shape parameters, and $\posecode$ as the concatenation of the inferred angles for a selected set of 13 joints in axis-angle representation corresponding to the joints situated above the `belly button' joint for a total of $\posecode=39$ elements. Selection of the joints ensures that joints that are typically not present in the TEDXPeople dataset, and thus cannot be reliably estimated, will not negatively affect the precision of the evaluation. We extract the yaw angle $\Theta$ by transforming the root joint axis angle rotation inferred by the model into the corresponding rotation matrix $M=M_y M_x M_z$, and extract the yaw angle $\Theta$ of the $M_y$ component representing the y-axis rotation matrix as follows:
\vspace{-0.2cm}
\begin{align*}
    \mathrm{pitch} &= \arcsin{M_{1,2}}\\
    \cos(\Theta) &= \frac{M_{2,2}}{\cos(\mathrm{pitch})}\\
    \sin(\Theta) &= \frac{M_{0,2}}{\cos(\mathrm{pitch})}\\
    \Theta &= \arctan_2(\sin(\Theta),\cos(\Theta)),
\end{align*}
\vspace{-0.05cm}
where the case of $\cos(\mathrm{pitch})=0$ is disregarded, since in practice we never render objects from high-pitch angles.

We now define the evaluation protocols followed for the animation and novel view synthesis tasks. For the animation task, we consider each test set video and select the first frame of each video as the source frame. We consider as driving frames five video frames, equally spaced along the duration of the test video. We then generate the object in the source frame in the pose of each driving frame under novel views, produced by rotating the object with the following $\Theta$ angles: $0, \pm\frac{\pi}{12}, \pm\frac{\pi}{6}, \pm\frac{\pi}{4}$. The triplets built from all combinations of source, driving and rendered frames are used for the computation of AYD, ASC and APC. For the novel view synthesis, we consider as source and driving frame the same, first frame of each video. This allows pure evaluation of the novel view synthesis capabilities of the method. We render each frame under the set of $256$ linearly sampled $\Theta$ angles in the range $[-\frac{\pi}{2},+\frac{\pi}{2}]$ and compute AYD, ASC and APC using all the available frame triplets.

\appprevspace
\section{Baseline details}
\label{ap:baselines}
\apppostvspace

\textbf{MRAA and FOMM.} MRAA and FOMM rely on affine transformations to transfer motion. Thus, in order to perform novel view synthesis a natural idea would be to modify these affine transformations such that they represent the object in the novel view. We achieve this with the following procedure. First, near each region center, we sample 4 additional keypoints in a small distance $=0.05$ forming a cross centered around the central keypoint for the region. The region center and these additional keypoints are then lifted to the 3D space using a depth map obtained from the driving image with off-the-shelf depth estimator~\cite{Omnidata}. As we need to recover depth for the object in the pose of the frame for which to perform novel view synthesis, we use absolute depth for animation to ensure the rendered frame pose is the same as the driving frame. These points are then projected to the target view using the desired camera parameters. Finally, we estimate a new affine transformation from these projected points. Note the off-the-shelf depth estimator only provides relative depth, and it is thus not possible to utilize it directly. To overcome this issue, we leverage depth obtained from our method and find a linear mapping between our depth and off-the-shelf depth. This linear mapping is consists of $d_{scale}$ and $d_{shift}$ parameters and can be find in closed form:
\vspace{-0.05cm}
\begin{equation*}
    d_{scale} = \frac{Cov(d, \hat{d})}{Var(\hat{d})},
\end{equation*}
\begin{equation*}
    d_{shift} = E\left[d\right] - d_{scale} E[\hat{d}],
\end{equation*}
\vspace{-0.05cm}
where $d$ is the depth map from our method, $\hat{d}$ is the off-the-shelf depth map, $E$ is sample mean, $Var$ is sample variance, and $Cov$ is sample covariance. 

\textbf{LIA.} LIA expresses animation as navigation inside a learned latent space. Given an embedding $z_s$ in this latent space for the source image, animation is expressed as $z_d = z_s + w = z_s + \sum{a_i d_i}$, with $z_d$ expressing the latent code corresponding to the animated result and vector $w$ expressed as the summation of a set of learned motion directions $d$ multiplied by corresponding magnitudes $a$, which form an orthogonal basis of the latent space. The set of learned motion directions represents the main types of motions performed by the objects. Interestingly, we find that for the VoxCeleb dataset, $d_2$, the second of such directions, is correlated with y-axis head rotation. While this movement is undesirably entangled with other motion components such as x-axis head rotation, we exploit this finding to produce novel views. Since no immediate correspondence between magnitude $a_2$ added to such direction and $\Theta$ exists a priori, we build a linear model mapping changes in $\Theta$ between the source and driving frame with $a_2$. To build such a linear model, we consider the first frame of each video and produce novel views using values of $a_2$ in the range of $[-17,+17]$ degrees. For each generated novel view, we evaluate the corresponding changes in $\Theta$ between the source and driving frame using 6DRepNet~\cite{hempel20226d} and use such data to fit our linear model $a_2=7.453 \Theta$. Given a desired $\Theta$ angle, we leverage the linear model to devise the magnitude $a_2$ and produce $z_\mathrm{novel} = z_d + a_2 d_2$, which is decoded to the frame under the novel view. Note that since the linear model directly optimizes the $\Theta$ error on the test set, we expect the AYD metric produced for such baseline to be biased toward lower values.
\appprevspace
\section{Novel view synthesis}
\label{ap:novelview-synthesis}
\apppostvspace

In this section, we evaluate the capabilities of the animation methods to perform novel view synthesis without animation. To this end, we simply rotate the image along the y axis on the set of angles from $-90^{\circ}$ to $+90^{\circ}$. The results are provided in Tab.~\ref{tab:novelview}, and confirm the findings from Sec.~\ref{sec:animation-eval}. While MRAA has favourable ASC and APC errors, it has very high AYD. This behavior is expected, because the method is simply performing a translation of the subject, rather than rotating it according to the provided yaw angles. This ensures the pose and identity remain preserved, at the cost of performing poor novel view synthesis. LIA, on the other hand, has a low AYD, while ASC and APC are high, which confirms that LIA has entangled latent directions that prevent novel view synthesis without significantly altering the pose and identity. Our model achieves the best AYD, which suggests that it performs the most accurate camera manipulations.

\begin{table}[t]
    \setlength{\tabcolsep}{5.0pt}
    \footnotesize
    \centering
    \begin{tabular}{c|ccc|ccc}
        \toprule
        &\multicolumn{3}{c}{VoxCeleb} & \multicolumn{3}{c}{TEDXPeople} \\
        \midrule
        Method & AYD$\downarrow$ & ASC$\downarrow$ & APC$\downarrow$ & AYD$\downarrow$ & ASC$\downarrow$ & APC$\downarrow$  \\
        \midrule
         FOMM \cite{FOMM} & 0.801 & 0.145 & 0.194 & 0.639 & 0.029 & 1.14 \\
         MRAA \cite{MRAA} & 0.760 & 0.133 & 0.177 & 0.686 & \textbf{0.022} & \textbf{0.861} \\
         LIA \cite{LatentImageAnimator}  & 0.188 & 0.132 & 0.198 & - & - & - \\
         Our 1 frame & 0.155 & \textbf{0.119} & \textbf{0.171} & 0.248 & 0.023 & 0.941 \\
         Our 5 frame & \textbf{0.153} & 0.126 & 0.184 & \textbf{0.244} & 0.024 & 0.959 \\
         \bottomrule
    \end{tabular}
    \vspace{-0.3cm}
    \caption{The results of generating novel views of the first frame of each video sequence. Camera angles range from $-90^{\circ}$ to $+90^{\circ}$.}
    \label{tab:novelview}
    \vspace{-0.5cm}
\end{table}

\appprevspace
\section{Canonical visualization}
\apppostvspace

\begin{figure}[t]
\centering
\begin{tabular}{cc}
     \includegraphics[width=0.45\linewidth]{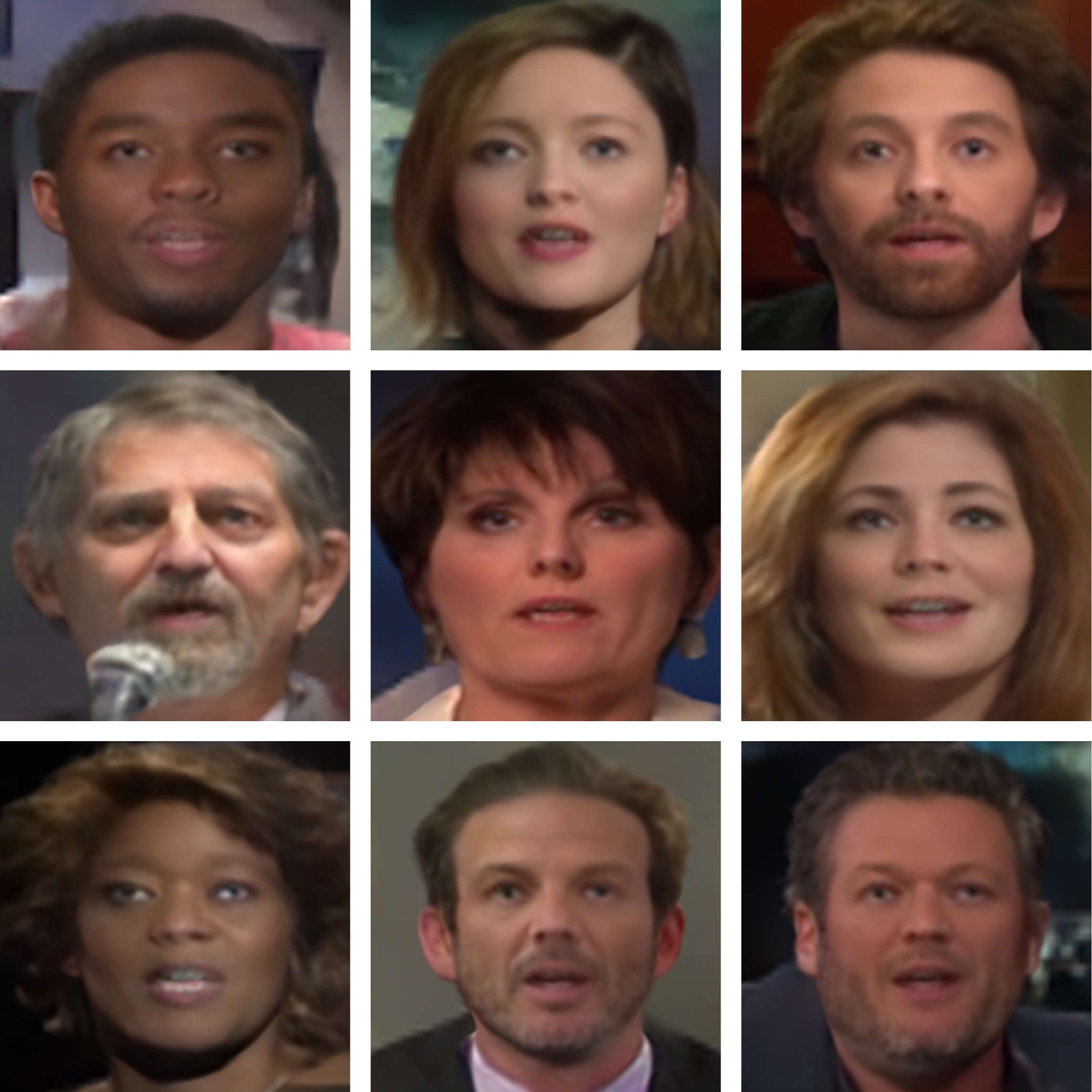} & \includegraphics[width=0.45\linewidth]{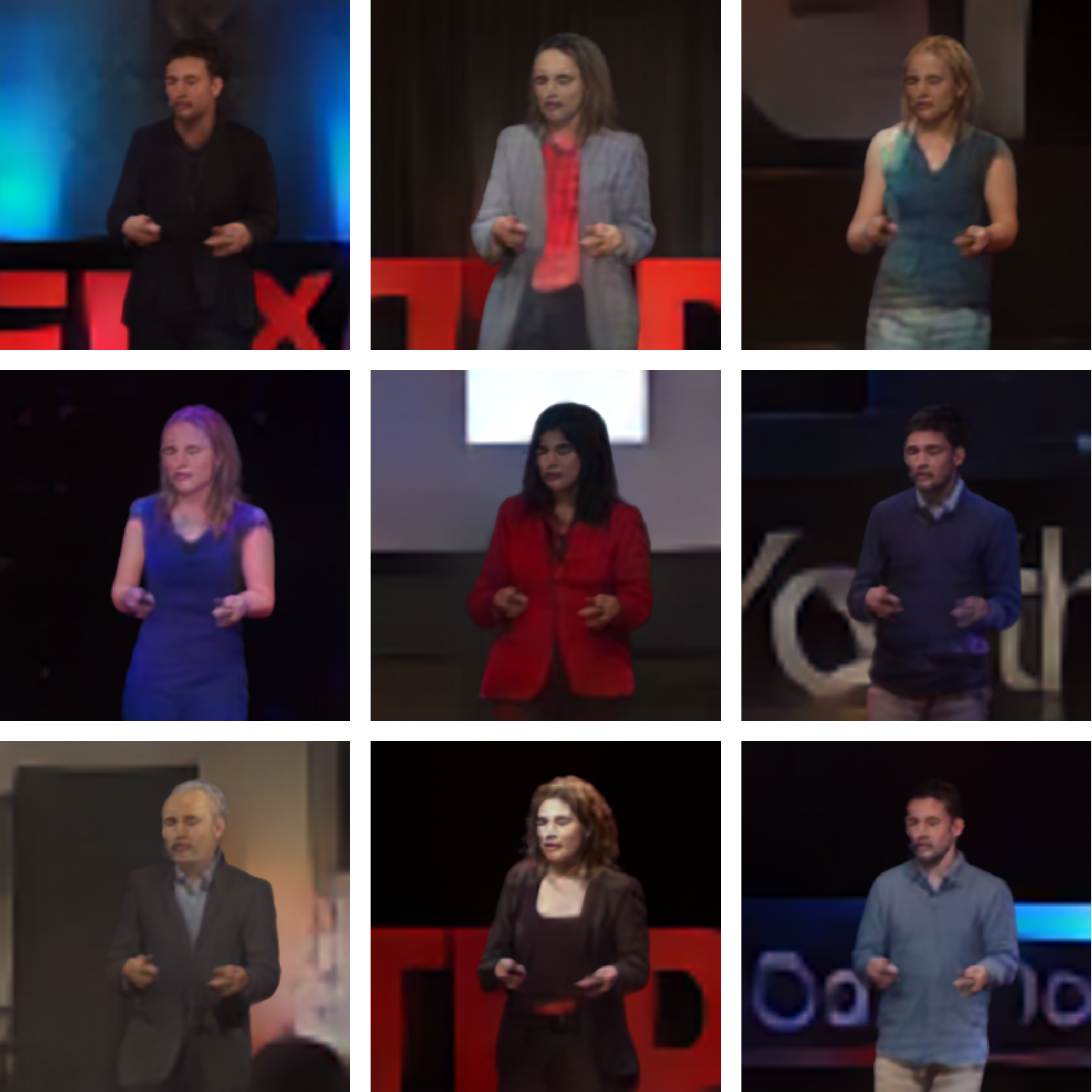}  \\
     (a) VoxCeleb~\cite{voxceleb} & (b) TEDXPeople~\cite{stylepeople}\\
\end{tabular}
\vspace{-0.3cm}
\caption{Visualization of canonical spaces.}
\label{fig:canonical}
\vspace{-0.5cm}
\end{figure}

In order to better demonstrate the representation learned by our model, we visualize some of the training identities in the canonical pose, i.e. where $\Tp$ is the identity matrix for all parts. Note that, since there are no prior assumptions on how the object should be placed in the rendering cube, parts seen from the camera with identity matrix extrinsics may be arbitrary. Thus, we select the camera from which objects will look reasonable. Note that $\Tp$ is still the same for all parts and all objects. The visual results are presented in Fig~\ref{fig:canonical}, which clearly shows that all objects have the same pose, which is a crucial property for animation.

\appprevspace
\section{Failed Experiments}
\label{ap:failed}
\apppostvspace

\textbf{Canonical space representation.} During our initial experiments we tried many different representations for canonical space: Triplanar~\cite{EG3D}, MLP~\cite{NeRF}, CP and VM decomposed cubes~\cite{TensoRF}. However, we found that decomposed solutions such as Triplanar~\cite{EG3D} and VM~\cite{TensoRF} are biased towards flat geometry, while an MLP~\cite{NeRF} is extremely slow. Note that the Triplanar~\cite{EG3D} representation also utilizes a small MLP, thus in our experiments it was slower than directly sampling our Voxel cube.

\textbf{Pose prediction.} Before reaching the PnP formulation, we tried many different approaches for pose prediction. First, we started with \textit{Direct} approaches, and we tested several architectures and rotation representations~\cite{zhou2019rotations}. However, all of them failed to produce meaningful geometry. We also tried an optimization-based approach for motion, i.e. having a pose parameter for each frame in the dataset. While this produced decent results for a single video, when the number of frames scales to millions, this approach quickly becomes infeasible.

\textbf{Different PnP.} We tested several different PnP implementations. We found that implementations based on declarative layers~\cite{gould2022deep, BPnP} are extremely slow, and using them in our setting would have been unfeasible. We also tested an implementation from the Kornia Library~\cite{kornia} that is based on DLT. However, it did not produce any meaningful results and produced divergence of the model. Our final choice was the EPnP~\cite{EPnP} implementation from Pytorch3D~\cite{pytorch3d}. However, we would like to note that it was only working in PyTorch 10.1 and not PyTorch 11, where it was not converging. We discovered the problem was the initially unstable gradients of the \textit{pinverse} function in the newer version. We think that this instability can be solved with better initialization for the 2d points, however we left this investigation for future work.

\textbf{Depth and normal supervision.} To help discovering the geometry we also tried to utilize depth and normal supervision from an off-the-shelf predictor~\cite{Omnidata}. Note that, because the normals supervision require computation of second order derivatives and we rely on voxel sampling with grid\_sample, we need a second derivative of grid\_sample, which is not implemented in PyTorch\footnote{https://github.com/pytorch/pytorch/issues/34704}. Thus, we develop a custom cuda kernel for the second derivative. While the depth and normal supervision helps to improve results for one-phase training, we found it to be unnecessary with two-phase training.

\textbf{Upsampler.} We render images in full resolution, however in prior experiments we utilize an upsampler. While this method works faster and consumes less memory, it produces less detailed geometry and worse view consistency.

\textbf{Different multipart representations.} We also tried two different representations for describing objects with multiple parts. The first had a shared radiance $\Vfeat$ volume, but a separate density for each part, while the second used different radiance and density volume for each part. Both of these strategies produce reasonable results. However, for them it is much harder to discover a large number of parts, and they usually degrade to solutions with only one or two parts being used.

\textbf{Few shot NeRF regularization.} We also tested several few-shot NeRF regularization techniques: entropy loss on the NeRF weights~\cite{kim2022infonerf}, loss on the weights from MiP NeRF~\cite{mip-NeRF}, surface normals regularization~\cite{verbin2022refnerf} and warping loss from MVCGAN~\cite{MVCGAN}. We found that all of them are unnecessary with our two phase training strategy.

\textbf{Discriminator.} To regularize the novel views, we also try to employ a Discriminator, similarly to 3D-GANs~\cite{EG3D, EpiGRAF}. In more detail, we first predict the pose of the object and then try to rotate this pose to generate the object in the novel view. This image is subsequently passed to the discriminator. However, we found it hard to find proper rotation ranges, thus the discriminator reduced the quality of the geometry in our experiments.

\appprevspace
\section{Ethical considerations}
\label{ap:ethical}
\apppostvspace

\textbf{Dataset usage.}
The primary datasets used in our experiments, VoxCeleb~\cite{voxceleb} and TEDXPeople~\cite{stylepeople}, contain publicly available videos of notable figures in public venues, \eg celebrities giving interviews and speakers giving presentations to large audiences.
These datasets have been released by and employed for prior academic research, \eg the works we use for our comparisons and evaluations.

Other datasets, such as Khan~\etal~\cite{KHAN2021479} and SURREAL~\cite{varol17_surreal} which are used for our ground-truth depth inference evaluations, contain realistic but synthetic images rendered from 3D models of human faces and bodies, respectively.
SURREAL~\cite{varol17_surreal} uses body scans and motion capture sequences generated from 3D capture of the appearances and performances of subjects who consented to have this captured and released for academic purposes.
Khan~\etal~\cite{KHAN2021479} contains facial images generated by perturbing characteristics such as facial identity, hair and clothing for models in a standard 3D modeling and rendering framework, and thus do not correspond to any particular person whose identity may be at risk of being revealed.
Each of these datasets are both publicly available and have been used for prior academic works.
As such, there are no particular concerns about violating the privacy or anonymity of our test subjects.

\textbf{Potential for bias in synthesis results.}
As with other data-driven methods for performance-driven animation, the amount of variation in characteristics such as gender, age, body type, and ethnicity that can be handled by our methods with the source and driving subjects while producing plausible synthesis results is dependent on the amount of such variations contained in the dataset.
While the variations in the real and synthetic images used in our experiments are limited by those in the aforementioned datasets used in our evaluations, \eg in typical celebrity videos and TEDx presentations, our method has no particular limitations towards such subjects, and thus could be deployed on other datasets containing different identity characteristics.
Deploying this approach in a manner which is fair and robust with respect to such variations for non-academic purposes, such as commercial applications, would require employing a dataset that is appropriately representative of the possible target identities, and evaluating the results to ensure consistent behavior across these demographics.
However, for the academic evaluations presented here, our evaluations suggest that our approach works as expected given the datasets we use, and thus could generalize to other training datasets fairly easily.
Finally, as our approach only relies on unconstrained video sequences for training, acquiring the data needed to adapt to new subjects is fairly straightforward, provided that the appropriate video sequences can be collected for training.
As such, there are no particular concerns related to unfair bias in our approach.

\textbf{Possibly misuse.}
As with other works in the domain of realistic, performance-driven animation, our work carries with it the possibility of use for deceptive activities, \eg, creating plausible videos of public figures as misinformation to advance a political agenda.
However, we maintain that, while this is clearly a valid concern for the near future, developing and studying such technology in public forms such as this work raises awareness of this potential, and with it the skepticism of viewers towards potentially misleading videos.
Furthermore, publicly describing our work and results allows for the advancement of forensic methods to identify when such manipulations have occurred.
We thus believe that our work helps to prevent the secretive development and deployment of these techniques for malicious ends which are not known or detectable either to average media consumers or professional forensic analysts.

\end{document}